\documentclass[runningheads]{llncs}
\usepackage[utf8]{inputenc}
\usepackage{graphicx}
\usepackage{hyperref}
\usepackage{amsmath}
\usepackage{hyperref}
\usepackage{caption}
\usepackage[table]{xcolor}
\usepackage{float}
\usepackage{paralist}
\usepackage{url}
\usepackage{tabto}
\usepackage{latexsym}
\usepackage{textcomp}
\usepackage{amsfonts}
\usepackage{hyphenat}
\usepackage{xcolor}
\usepackage{amssymb}
\usepackage{algorithm}
\usepackage[noend]{algpseudocode}
\usepackage{listings}
\usepackage{subcaption}
\usepackage{paralist}
\usepackage{url}
\usepackage{comment}
\usepackage{proof}
\usepackage{colortbl}
\usepackage{multirow}
\definecolor{powderBlue}{RGB}{147, 207, 207}
\definecolor{blueGreen}{RGB}{89, 189, 192}
\definecolor{metallicSeaweed}{RGB}{6, 134, 146}
\definecolor{viridianGreen}{RGB}{1, 146, 150}
\definecolor{burntSienna}{RGB}{220, 120, 85}
\definecolor{melon}{RGB}{246, 183, 161}
\definecolor{rust}{RGB}{190, 51, 2}
\definecolor{grey}{RGB}{183, 183, 183}

\newcommand{\furl}[1]{\footnote{\scriptsize \url{#1}}}

\begin{document}
\title{\textit{EABlock}: A Declarative Entity Alignment Block for Knowledge Graph Creation Pipelines}
\titlerunning{\textit{EABlock}: A Declarative Entity Alignment Block for KG Creation Pipelines}

\author{Samaneh Jozashoori\inst{1,2}\orcidID{0000-0003-1702-8707},
        Ahmad Sakor \inst{1,2}\orcidID{0000-0001-8007-7021},
        Enrique Iglesias\inst{3}\orcidID{0000-0002-8734-3123},
        Maria-Esther Vidal\inst{1,2,3}\orcidID{0000-0003-1160-8727}
        }
%
\authorrunning{Jozashoori et al.}
%
\institute{TIB Leibniz Information Center for Science and Technology, Germany\\
\email{\{samaneh.jozashoori,ahmad.sakor,maria.vidal\}@tib.eu}
\and
Leibniz University of Hannover, Germany
\and
L3S Research Center, Germany\\
\email{iglesias@l3s.de}
}
\maketitle        
\begin{abstract}
Despite encoding enormous amount of rich and valuable data, existing data sources are mostly created independently, being a significant challenge to their integration.
Mapping languages, e.g., RML and R2RML, facilitate declarative specification of the process of applying meta-data and integrating data into a knowledge graph. Mapping rules can also include knowledge extraction functions in addition to expressing  correspondences among data sources and a unified schema. Combining mapping rules and functions represents a powerful formalism to specify pipelines for integrating data into a knowledge graph transparently. Surprisingly, these formalisms are not fully adapted, and many knowledge graphs are created by executing \emph{ad-hoc} programs to pre-process and integrate data. In this paper, we present \textit{EABlock}, an approach integrating Entity Alignment (EA) as part of RML mapping rules. \textit{EABlock} includes a block of functions performing entity recognition from textual attributes and link the recognized entities to the corresponding resources in Wikidata, DBpedia, and domain specific thesaurus, e.g., UMLS. \textit{EABlock} provides agnostic and efficient techniques to evaluate the functions and transfer the mappings to facilitate its application in any RML-compliant engine. We have empirically evaluated \textit{EABlock} performance, and results indicate that \textit{EABlock} speeds up knowledge graph creation pipelines that require entity recognition and linking in state-of-the-art RML-compliant engines. \textit{EABlock} is also publicly available as a tool through a \href{https://github.com/SDM-TIB/EABlock}{GitHub repository} and a \href{https://doi.org/10.5281/zenodo.5779773}{DOI}.
\end{abstract}

\begin{figure}[H]
\centering
\includegraphics[width=\linewidth]{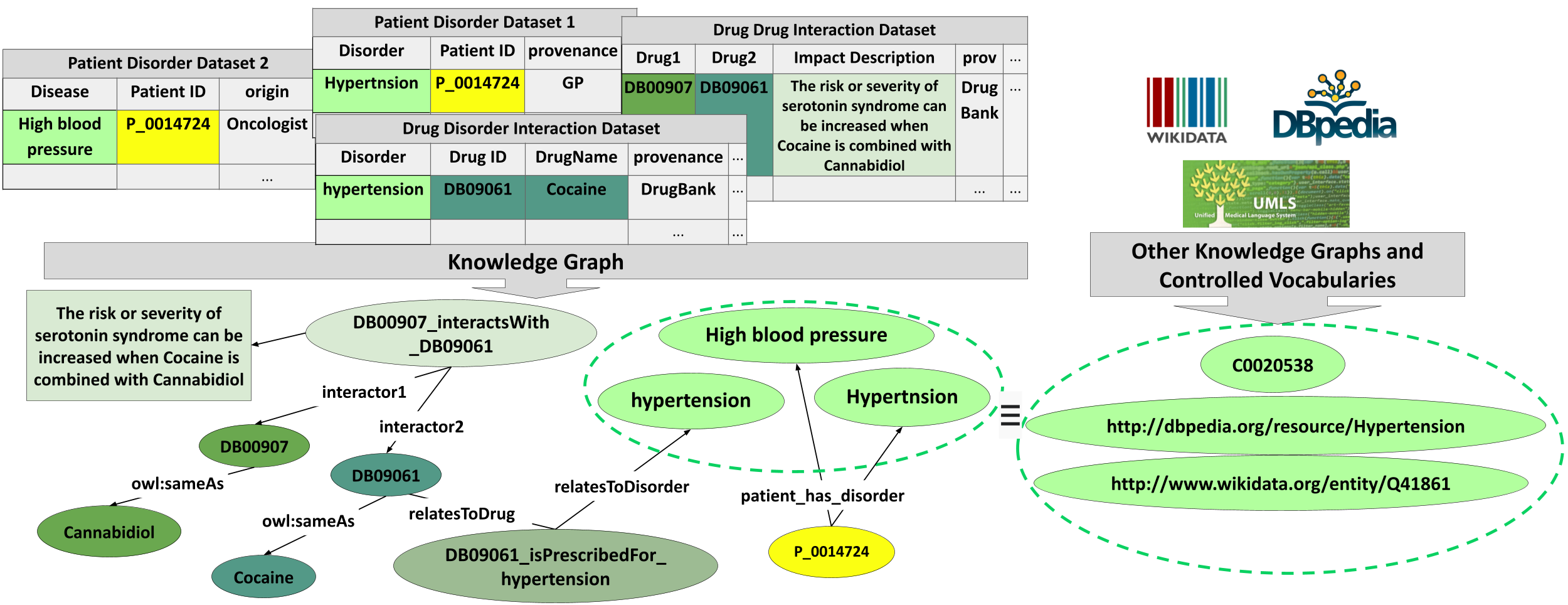}
\caption{\textbf{Motivating Example}. Data integration from four datasets. Different interoperability issues: concept disorder is modeled differently in Dataset 1 and 2. The entity hypertension is represented with various entities, and its name is misspelled in Dataset 1. Performing EA with UMLS, DBpedia, or Wikidata, enables conflict resolution and integration into a KG. }
\label{fig:motivation}
\end{figure}
\section{Introduction}
Knowledge graphs (KGs) represent the convergence among data and knowledge using networks. Albeit coined by the research community for several decades, KGs are playing an increasingly relevant role in scientific and industrial areas~\cite{GutierrezS21}.
Years of research on semantic data management and knowledge engineering have paved the way for the integration of factual statements spread across various data sources or collected from community-maintained data sources (e.g., Wikidata~\cite{DBLP:journals/cacm/VrandecicK14} and DBpedia~\cite{DBLP:conf/semweb/AuerBKLCI07}). 
The rich spectrum of knowledge represented in existing KGs, position them 
as sources of background knowledge to empower data-driven processes. 
Nevertheless, real-world applications require accountable methods to facilitate the traceability of data management processes performed to integrate data into a KG.
Thus, KG management needs to be enriched with transparent methods to understand and validate the steps performed to transform disparate data into a unified KG.

\noindent Data integration systems (DIS)~\cite{lenzerini2002data} represent generic frameworks to define a KG in terms of a unified schema, a set of data sources, and mapping rules between concepts in the unified schema and the sources. The declarative definition of mapping languages represents a building block for tracking down a KG creation; it also facilitates reusability and modularity. Mapping languages (e.g., R2RML~\cite{das2012r2rml} and RML~\cite{dimou2014rml}) have been proposed as standards describing correspondences between the concepts in the unified schema (e.g., classes, properties, and relations) and the data sources' attributes. Thus, by following the global as view (GAV) paradigm~\cite{lenzerini2002data} where concepts in the unified schema are defined in terms of the sources, they enable the resolution of interoperability conflicts among data sources defined using different schemas. However, data sources may have diverse levels of structuredness (e.g., structured, semi-structured, and unstructured), suffer from data quality issues, or present several interpretations of the same real-world entity. The resolution of these conflicts as part of the process of KG creation can be defined as Data Operators in a Data Ecosystem (DE) proposed by Capiello et al.~\cite{capiello2020data}. Alternatively, mapping languages have been extended to embrace Data Operators as functions that can be included as programming scripts directly in the mapping rules~\cite{debruyne2016r2rml,junior2016funul,vu2019d} or can follow a declarative approach (e.g., using the Function Ontology, FnO)~\cite{de2016ontology}. They offer clear benefits in comparison to ad-hoc pre- and post-processing techniques in terms of reusability and reproducibility. Nonetheless, the lack of generic frameworks to deal with mapping rules and functions complicates mapping rule design because these functions need to be also implemented.

\noindent\textbf{Our Method:} 
We address the problem of EA using target knowledge to solve interoperability conflicts across data sources by proposing a method named \textit{EABlock}. \textit{EABlock} is a computational block composed of a set of FnO functions which can be called from RML mapping rules and an efficient strategy to evaluate them. The functions in \textit{EABlock} are tuned to effectively align entities in a KG with their corresponding entities in existing KGs (e.g., DBpedia~\cite{DBLP:conf/semweb/AuerBKLCI07} and Wikidata~\cite{DBLP:journals/cacm/VrandecicK14}) and controlled vocabularies (e.g. UMLS~\cite{UMLS}). These functions resort to another engine for solving the tasks of name entity recognition (NER) and entity linking (EL) required for EA; any engine performing NER and EL tasks can be utilized.  
\textit{EABlock} follows an \emph{eager evaluation} strategy and enables the execution of the \textit{EABlock} functions before the RML mapping rules are executed. This evaluation strategy defined by Jozashoori et al.~\cite{jozashoori2020funmap}, facilitates the transformation of RML with the \textit{EABlock} functions into function-free RML mapping rules that can be executed by any RML-compliant engine without requiring any modification in the engine. \textit{EABlock} has been developed and experimentally evaluated in real-world datasets collected from DBpedia,  Wikidata, and UMLS. The observed outcomes suggest that \textit{EABlock} functions perform EA to domain-specific and encyclopedic KGs effectively. \textit{EABlock} is utilized in three international projects to integrate data into the KGs developed in these projects. The results corroborate the role that declaratively defined functions have in KG management. \\
\noindent
This paper is structured in six additional sections. Section \ref{sec:rw} summarizes the state of the art, and section \ref{sec:example} motivates and defines the problem addressed by \textit{EABlock}. While section \ref{sec:pre} provides an overview of background knowledge and concepts, section \ref{sec:approach} formally defines the problem and describes \textit{EABlock} as the solution including its proposed strategy and techniques. In section \ref{sec:eval} the results of the experimental study are reported. Lastly, section \ref{sec:conclusion} wraps up and outlines future work.    

\raggedbottom

\section{Related Work}
\label{sec:rw}
Entity Alignment (EA) is an important solution to overcome interoperability issues while creating a knowledge graph from heterogeneous data sources. Dimou et al.~\cite{dimou2017ilastic}, Michel et al.~\cite{michel2020covid}, and Vidal et al.~\cite{vidal2019transforming} propose EA as a pre-processing step, prior to the semantic enrichment and integration of data. In this case, pre-processing performs the task of EA on the whole provided data sources, independent of their involvement in the goal KG. Hence, including EA as part of pre-processing can add a considerable overhead on the knowledge graph creation pipeline. Additionally, pre-processing steps are usually developed as \textit{ad-hoc} programs, which are neither declarative nor easy to maintain. SemTab\furl{https://www.cs.ox.ac.uk/isg/challenges/sem-tab/} is an effort in benchmarking systems dealing with the tabular data to KG matching problem and present existing challenges~\cite{jimenez2020results}. An alternative is to perform EA after the creation of KG, at the expense of creating the same nodes multiple times across different KGs. Zeng~\cite{zeng2021comprehensive} provides a comprehensive survey of available techniques to add EA in post-processing, to find the equivalent entities in different created KGs. Lastly, EA can be part of the main pipeline of semantic data integration and knowledge graph creation applying transformation functions. In other words, EA can be involved in the mapping rules that enrich raw data semantically and transform them into RDF model. In this case, EA needs to be defined as a transformation function in the mapping rules. There are different mapping languages enabling the involvement of functions as part of the mapping rules such as RML+FnO~\cite{de2016ontology}, R2RML-F~\cite{debruyne2016r2rml}, FunUL~\cite{junior2016funul}, and D-REPR~\cite{vu2019d}. There also exist different engines capable of processing functions in different languages. For instance, FunMap~\cite{jozashoori2020funmap} is able of interpreting function-based mappings in RML+FnO into equivalent function-free mappings in RML efficiently. In spite of all the value that declarative mapping languages and corresponding techniques provide, their potential applications in the task of EA are neither well explored nor appreciated. Hereupon, we aim to fill this gap by enabling and facilitating the application of EA tools as part of mapping rules using transformation functions.

\begin{figure}[H]
\centering
\includegraphics[width=\linewidth]{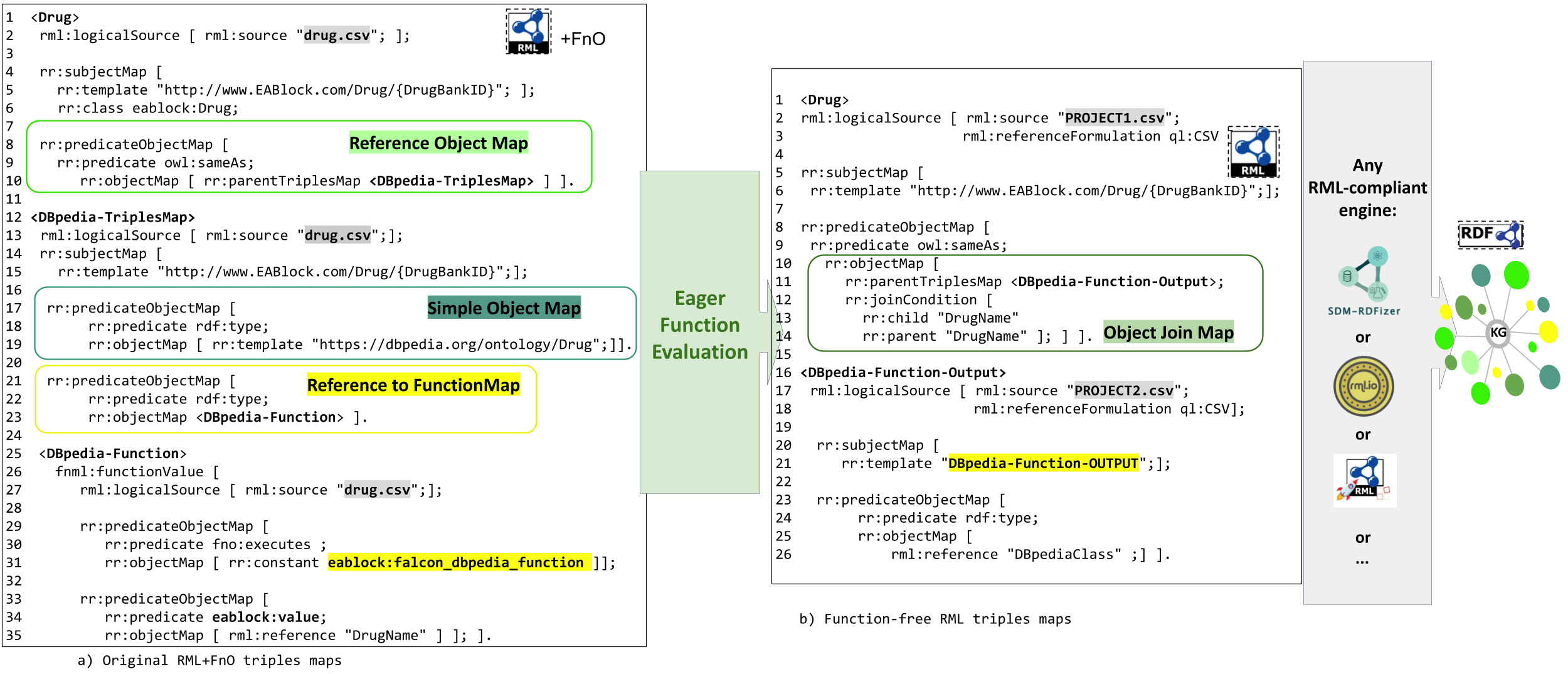}
\vspace{-0.5cm}
\caption{\textbf{RML+FnO Triples Maps}. a) \texttt{Drug} and \texttt{DBpedia-TriplesMap} are RML triples maps (lines 1-23), while \texttt{DBpedia-Function} is a FnO function (lines 25-35). b) Eager evaluation of FnO functions creates \texttt{PROJECT1.csv} and \texttt{PROJECT2.csv}, and generates function-free RML maps, which can be executed in any RML-compliant engine. Function configuration is not required.}
\label{fig:preliminaries}
\end{figure}
\raggedbottom
\section{Motivating Example}
\label{sec:example}
We motivate our work with a mock example from a real-world scenario illustrated in \autoref{fig:motivation}. In this scenario, the aim is to integrate four datasets obtained from different sources into a KG. The datasets consist of 
\begin{inparaenum}[\bf a\upshape)]
    \item Patient data extracted from two different clinical notes provided by a general practitioner (GP) and an oncologist including the comorbidities from which the patient is suffering, and
    \item  The drug related data extracted from DrugBank\furl{https://go.drugbank.com/} including drug-drug-interaction providing information on the possible interactions between different drugs and the impact on effectiveness of each, and drug-disorder data revealing information on list of drugs that can be prescribed for each disorder. 
    \end{inparaenum}
    A portion of the KG created by a naive approach can be observed in \autoref{fig:motivation}. A closer look reveals that the same disorder instance exists as three separated nodes in the graph, i.e., there is an interoperability conflict among them. The existing interoperability issue can be traced back to the raw data where \textbf{I.} the same disorder is represented with different names by clinical physiologists, and \textbf{II.} the name of the disorder is misspelled in one of the records. Another important point is regarding the connection between the instances of the generated KG and instances in available domain-specific sources (e.g., UMLS) or encyclopedic KGs (e.g., DBpedia and Wikidata) which represent the same real-world entities. More specifically, the importance of mentioned connections appears while integrating or linking the other available data/knowledge bases, which are annotated by instances of such sources (i.e., UMLS, DBpedia, and Wikidata). Both observations emphasize the importance of including EA as a module in the pipeline of KG creation. It should be noted that following FAIR principles~\cite{wilkinson2016fair}, transparency and reproducibility are essential requirements in pipelines of KG creation. All blocks applied as part of the main process or pre- or post-processing of KG creation should be transparent and traceable. This leads to thinking about an independent transparent module for entity alignment, using a declarative language that can be integrated in any KG creation pipeline that is compliant with the same mapping language.

\section{Preliminaries}
\label{sec:pre}

Knowledge graphs (KGs) are data structures that represent factual knowledge as entities and their relationships using a graph data model~\cite{GutierrezS21}. A KG is a directed graph $G$=($O$,$V$,$E$), where:
\begin{inparadesc}
    \item[$O$] is a unified schema that comprises classes, properties, and relations. 
    \item[$V$] is a set of nodes in the KG; nodes in V correspond to classes or instances of classes in O. 
    \item[$E$] is a set of directed labeled edges in the KG that relate nodes in $V$. Edges are labeled with properties and relations in $O$. 
\end{inparadesc}
A KG creation process can be specified in terms of a Data Ecosystem (DE). A DE~\cite{capiello2020data} is defined as a 4-tuple $DE=\langle Data Sets, Data Operators, Meta$-$Data, Mappings\rangle$ where $Data Operators$ represent a set of operators that can be executed over data in $Data sets$, including a set of structured or unstructured data sets. $Meta$-$Data$ describes the domain of knowledge and meaning of the data residing in $Data Sets$ accordingly. $Meta$-$Data$ comprises: 
\begin{inparaenum}[\bf I \upshape.]
\item Ontologies and controlled vocabularies to provide a unified view of the domain knowledge;
\item Properties to describe the data quality, provenance, and access regulations; and 
\item descriptions of the main characteristics of data.
\end{inparaenum}
Finally, $Mappings$ represent the correspondences among the concepts and properties in different domain ontologies or associations between data in $Data Sets$ and the domain ontology. 
\noindent The same real-world entity can be represented differently in the data sources in $Data Sets$. Interoperability issues include: 
\begin{inparadesc}
\item[\emph{Structuredness}:] this conflict occurs whenever data sources are described at different levels of structuredness, e.g., structured, semi-structured, and unstructured. \item[\emph{Schematic}:] this interoperability conflict exists among data sources that are modeled using different schemas, e.g., different attributes representing the same concept.
\item[\emph{Domain}:] this interoperability conflict occurs among various interpretations of the same entity. They include: i) homonym: the same name is used to represent concepts with different meaning, and ii) synonym: distinct names are used to model the same concept.
\end{inparadesc}
\autoref{fig:motivation} illustrates the interoperability issues: structuredness between the two data sources of drug-drug interactions; schematic among the attributes of dataset 1 and 2; and domain among the names representing the hypertension. In general, KG creation pipelines include an additional pre- / post-processing block to solve the interoperability issues between data sources. However, this block can be part of the DE as a $DataOperators$ utilizing the knowledge encoded in the $Meta$-$Data$~\cite{abs-2105-09312}.

\noindent KGs are expressed in the Resource Description Framework (RDF), where nodes  can be resources or literals, and edges correspond to predicates. RDF resources are identified by IRIs (Internationalized Resource Identifier) or blank nodes (anonymous resources or existential variables), while literals correspond  to instances of a data type (e.g., numbers, strings, or dates).
Mapping rules in $DE$ are declaratively defined using the RDF Mapping Language (RML), an extension of the W3C-standard mapping language R2RML. RML allows for the definition of sources in different formats (e.g., CSV, Relational, JSON, and XML). An RML mapping rule, named \texttt{TriplesMap}, follows the global as view paradigm~\cite{lenzerini2002data}, i.e., concepts in the unified schema are defined in terms of a data source. \autoref{fig:preliminaries} presents RML \texttt{TriplesMap}s. A \texttt{rr:subjectMap} defines the resources of an RDF class in the unified schema, while
a set of predicate-object maps (\texttt{rr:predicateObjectMap}) define the properties and relations of a class. 
The values of a predicate-object map can be defined in terms of a data source attribute, or as a reference or a join with the \texttt{rr:subjectMap} in another \texttt{TriplesMap}. A reference to another triples map is denoted as \texttt{rr:RefObjectMap}; it can be stated only between triples maps defined over the same data source. Lastly, a \texttt{rr:JoinCondition} represents references between \texttt{TriplesMap} defined on different data sources. A function can define \texttt{rr:subjectMap} or \texttt{rr:predicat}-\texttt{eObjectMap}. The Function Ontology (FnO) is used to specify functions of the type \texttt{FunctionMap}~\cite{de2016ontology}. 
A $DE$ where its $Mappings$ comprises RML \texttt{TriplesMap}s with FnO functions, can be executed following two strategies.
\begin{inparaenum}[ a\upshape)]
\item \textit{Lazy evaluation} delays the execution of a function until when it is needed to compute a value in a \texttt{TriplesMap}. 
\item\textit{Eager evaluation} executes the functions in $Mappings$ over the data sources before these values are needed in the RML triples maps. 
\end{inparaenum}
The lazy evaluation requires an understanding of RML and FnO. Contrary, the eager evaluation enables the transformation of the RML+FnO \texttt{TriplesMap}s into function-free RML triples maps. This evaluation can be done beforehand, and the results can be represented as sources of the translated function-free \texttt{TriplesMap}s. Another advantage of an eager evaluation is that an RML-compliant engine can be used to execute the function-free \texttt{TriplesMap}s and create a KG. \autoref{fig:preliminaries} a) presents two RML \texttt{TriplesMap}s (lines 1-23) and the function \texttt{DBpedia-Function} is defined in lines 25-35. Following a lazy evaluation,  \texttt{DBpedia-Function} is executed each time a new  entity of the class Drug is created.   
This execution requires that the RML engine is able to execute functions. Also, in presence of large number of duplicates in the data sources (i.e., \texttt{drug.csv}), it may be executed several times. 
On the other hand, \autoref{fig:preliminaries} b) depicts the translation performed for eager evaluation; this approach is described by Jozashoori et al. ~\cite{jozashoori2020funmap}. The transformation RML \texttt{TriplesMap}s are evaluated over new data sources. The data source \texttt{PROJECT1.csv} is created from the \texttt{drug.csv} following well-known properties of the relational algebra (e.g., pushing down of projections and selections into the data sources); they enable not only the reduction of the size of data sources but also the elimination of duplicates. Moreover, \texttt{PROJECT2.csv} is created from the materialization of \texttt{DBpedia-Function}. The reference between the two \texttt{TriplesMap}s is expressed with a \textit{join condition}.

\begin{figure}[H]
\centering
\includegraphics[width=\linewidth]{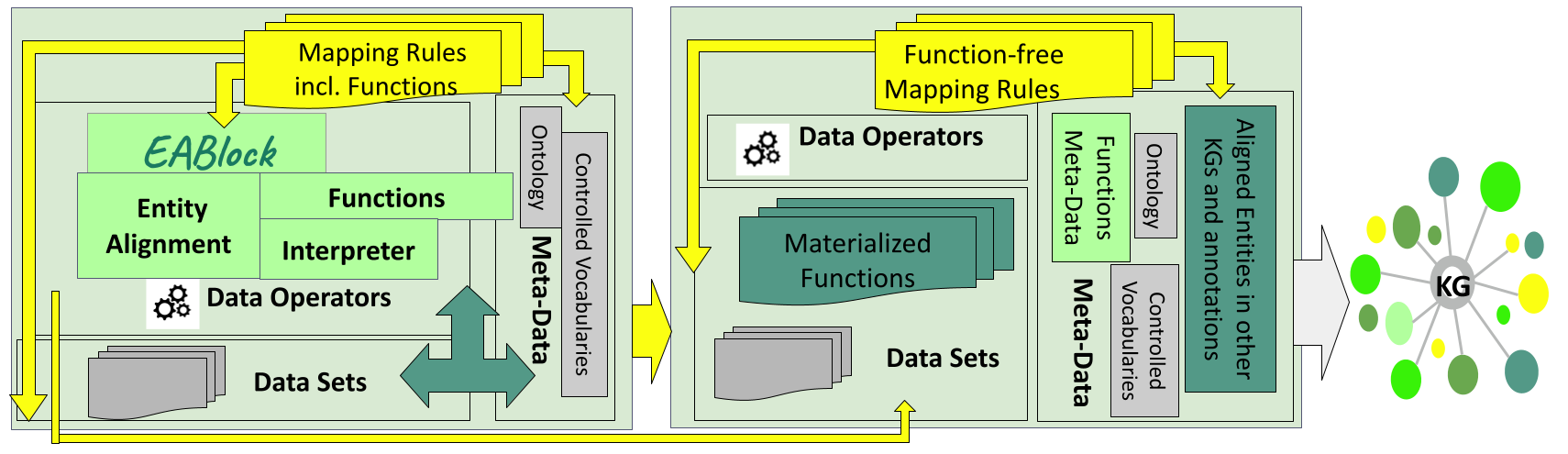}
\vspace{-0.25cm}
\caption{ The \textit{EABlock} components. A set of FnO functions that resorts to an Entity Alignment engine. An Interpreter that executes \textit{EABlock} functions included in  RML+FnO mapping rules and translate these rules into function-free rules.}
\label{fig:architecture}
\end{figure}

\section{Our Approach: \textit{EABlock}}
\label{sec:approach}
\noindent\textbf{Problem Statement:} As shown in \autoref{fig:motivation}, a KG can comprise entities that correspond to the same real-world entity (e.g., various entities representing hypertension). 
We address the problem of aligning entities in a KG $G_1$=($O_1$,$V_1$,$E_1$) with entities in an existing KG $G_2$=($O_2$,$V_2$,$E_2$) efficiently. Encyclopedic KGs like DBpedia~\cite{DBLP:conf/semweb/AuerBKLCI07} or Wikidata~\cite{DBLP:journals/cacm/VrandecicK14}, or domain-specific (e.g., UMLS~\cite{UMLS}) correspond to KGs $G_2$ against where the alignment is performed. 
\noindent\textbf{Proposed Solution:} Entity alignment from $G_1$ to $G_2$, $\gamma(G_1\mid G_2)$, is defined in terms of an \emph{ideal KG}, $G^*=(O^*,V^*,E^*)$, that includes the nodes and edges in $G_1$ and $G_2$ plus all the edges that relate nodes in $G_1$ with nodes in $G_2$. A solution to $\gamma(G_1\mid G_2)$ corresponds to a \emph{maximal partial function} $\zeta$:$V_1 \rightarrow V_2$ such that $\gamma(G_1\mid G_2,\zeta)$=$\{(s_1, sameAs, \zeta(s1))\mid (s_1, sameAs, \zeta(s1)) \in E^*\}$\footnote{A partial function $\zeta$:$V_1 \rightarrow V_2$ is a function from a subset of the $V_1$. $\zeta$ is maximal in the partial ordered set of all the functions from $V_1 \rightarrow V_2$.}. $DE_{G_{1,2}}=\langle Data Sets_{1}, Data Operators, Meta$-$Data_{1}, Mappings_{1,2}\rangle$ defines the KG, $G_{1,2}$=$(O_1 \cup \{sameAs\}, V_1\cup V_2, E_1 \cup \gamma(G_1\mid G_2,\zeta))$. The set $Mappings_{1,2}$ is a superset of $Mappings_{1}$ including all triples maps that define $\zeta$ and enable the computation of $\gamma(G_1\mid G_2,\zeta)$.
\raggedbottom

\noindent\textbf{\textit{EABlock}} is an approach proposing a computational block to solve entity alignment over textual attributes providing techniques bridging and utilizing all components of a DE i.e., $DataSets$, $Data Operat$- $or$, $Meta$-$Data$, $Mappings$: 
\begin{inparaenum}[\bf a\upshape)]
\item \textit{EABlock} links entities encoded in labels and short text to controlled vocabularies described by meta-data and resources in encyclopedic and other domain-specific KGs. For this purpose, \textit{EABlock} introduces a set of operating functions resorting to an entity and relation linking tool.  
\item \textit{EABlock} functions are defined in a human and machine-readable medium, meeting the requirements of meta-data in terms of transparency and reusability. Although the outcome of \textit{EABlock} representing the aligned entities and annotations provides meta-data for the KG, the addition of \textit{EABlock} functions to the meta-data of the DE equips this layer for further reproduction or maintenance of the KG with newly added data. 
\item \textit{EABlock} functions can be easily integrated into the mappings expressing the relations among the data and the ontology using RML language, applying available extensions of the language.
\item \textit{EABlock} also provides an efficient evaluation strategy to materialize the calls of the functions in the mappings extending data sources and transforming mappings to function-free RML mappings that are adaptable by any RML-compliant KG creation pipeline.   
\end{inparaenum}

\noindent As shown in \autoref{fig:architecture}, \textit{EABlock} composes three components: 
\begin{inparadesc}
\item[Functions] including the signatures of the \textit{EABlock} functions in FnO. 
The functions can be divided into two categories based on their domains and ranges.
\textbf{Keyword-based} functions receive case-insensitive keywords as input and generate one entity as the output, and \textbf{Short text-based} functions accept a case-insensitive short text as input and output a list of entities. \item[Entity Alignment] performs the NER and EL tasks. This component is agnostic, i.e., any tool solving the tasks of Named Entity Recognition (NER) and Entity Linking (EL) through an API can be employed; as a proof of concept, we use Falcon2.0~\cite{Sakor2020}.
\item[Interpreter] connects the previous two components. It follows an eager evaluation strategy of the functions and retrieves the results of the entity alignment generated by the entity alignment tool. The eager evaluation strategy gives the basis for an efficient and RML engine-agnostic execution of the \textit{EABlock} functions. It resorts to the approach proposed by Jozashoori et al.~\cite{jozashoori2020funmap} to translate the input RML+FnO \texttt{TriplesMap}s into function-free RML \texttt{TriplesMap}s. As explained before, \textit{EABlock} creates a new data set- output dataset- materializing the functions. The output dataset comprises two attributes; input and output attributes (\textit{attr1} and \textit{attr2}). Depending on the category of the function, \textit{EABlock} deploys one of the following two techniques. \textbf{a.} If the function is a \textbf{Keyword-based} function, for each input value, one record is added to the output dataset including the input value and the retrieved linked entity as the values of \textit{attr1} and \textit{attr2}, respectively. \textbf{b.} However, if the function is \textbf{Short text-based}, after evaluation of the function and receiving the list of linked entities, \textit{EABlock} generates the output dataset including one record for each entity in the list of linked entities, i.e., for each entity in the list of the retrieved linked entities, one record is added to the output dataset which includes input value and the linked entity as the values of \textit{attr1} and \textit{attr2}, respectively. In this way, \textit{EABlock} ensures that the generated datasets can be translated by any RML-compliant engine and result in exactly the same RDF triples; since different RML engines may have different interpretations of an RDF list.   
\end{inparadesc}
\noindent\textbf{Implementation and Application}
\textit{EABlock} approach is implemented and available as a tool. As a proof of concept, \textit{EABlock} integrates Falcon2.0 API \furl{https://labs.tib.eu/sdm/falconmedical/falcon2/} to perform the NER and EL tasks.
Falcon~\cite{SakorMSSV0A19,Sakor2020} is empowered with background knowledge that allows for the accurate recognition and linking of biomedical concepts. \textit{EABlock} is developed in Python3, open-source, and available under the Apache License 2.0~\footnote{To keep the anonymity of the paper, \textit{EABlock} will be publicly available as a tool at the Zenodo platform, GitHub, and Dockerhub \textbf{after} the review procedure.}. 

\begin{figure}[H]
    \centering 
    \subfloat[The performance of a KG creation pipeline applying RocketRML.]{
        \includegraphics[trim=0 0 0 21,clip,width=0.5\linewidth]{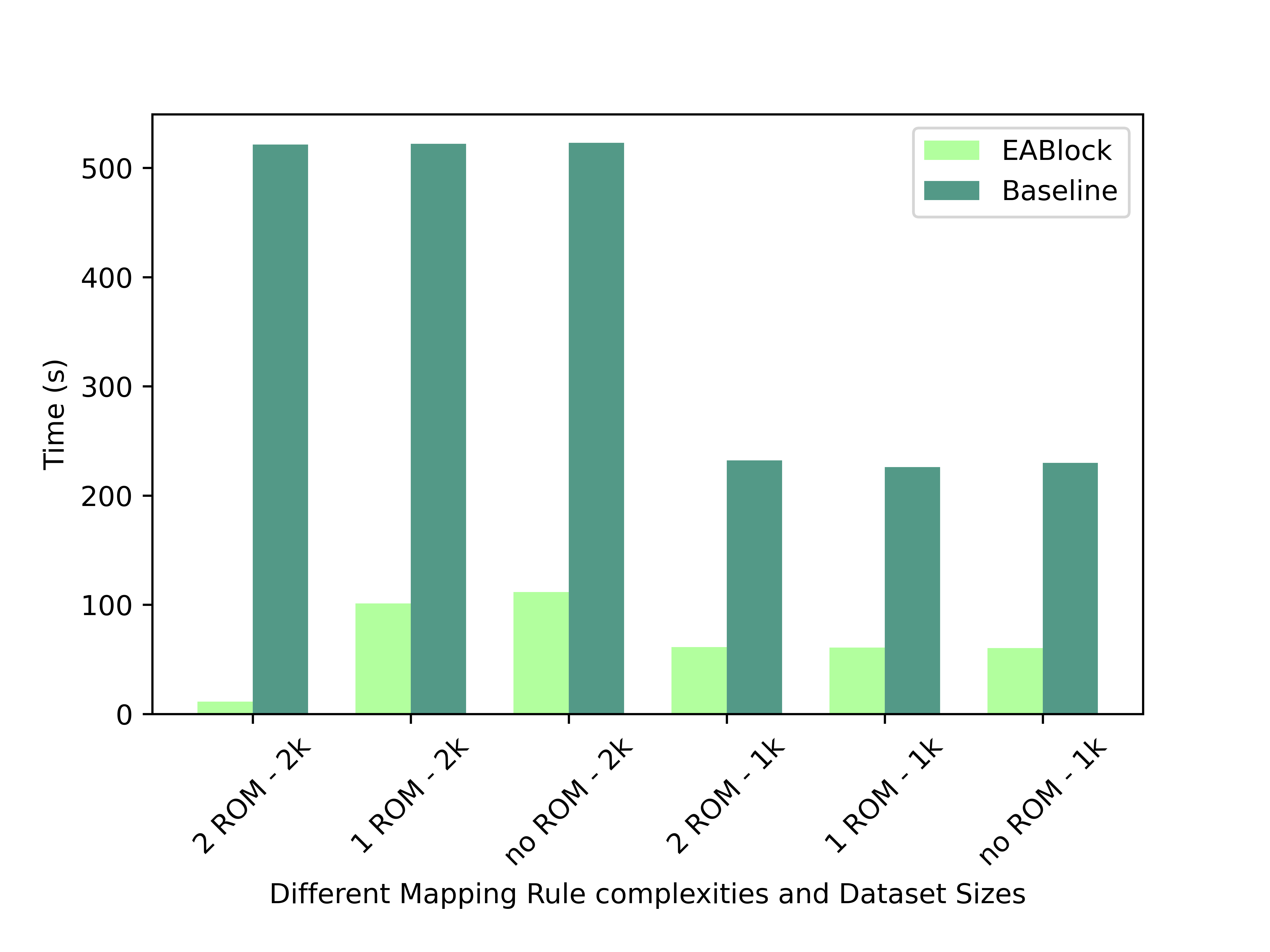}
        \label{fig:rocket}
    }
    \subfloat[The performance of a KG creation pipeline applying SDM-RDFizer.]{
        \includegraphics[trim=0 0 0 21,clip,width=0.5\linewidth]{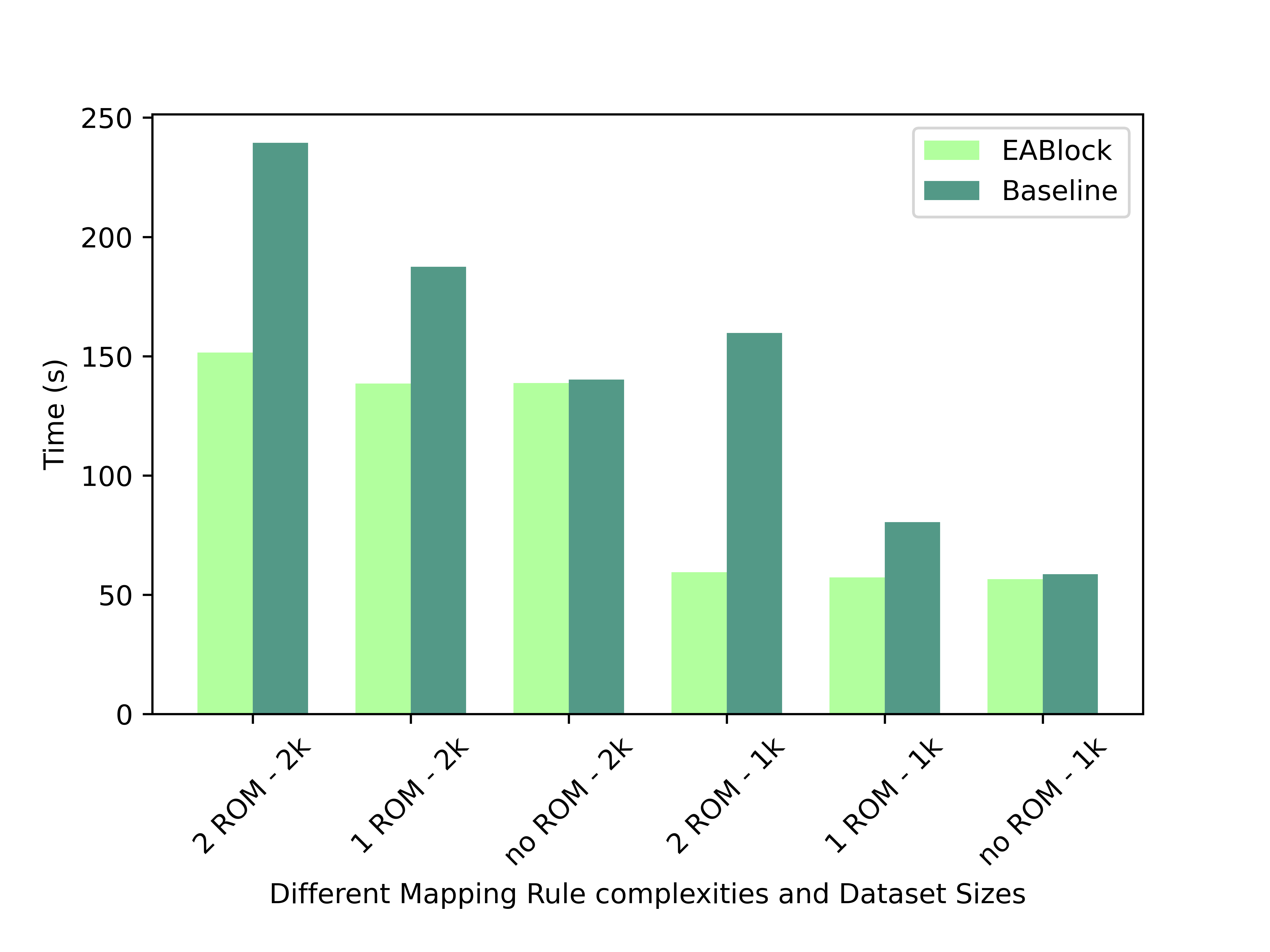}
        \label{fig:rdfizer}
    }
    \caption{\textbf{Efficiency}. The impact of using \textit{EABlock} in KG creation pipelines applying two different RML-compliant engines. Baseline corresponds to the execution of entity alignment in a pre-processing stage, while \textit{EABlock} enables the specification of this process in the RML mapping rules. As observed, \textit{EABlock} reduces the execution time of KG creation pipelines that involve entity alignment tasks in comparison to the application of the same functions but during a pre-processing stage.}
    \label{fig:efficiency}
\end{figure}

\section{Empirical Evaluation}
\label{sec:eval}
Our goal is to empirically assess the performance of the \textit{EABlock} in the resolution of the problem presented in section \ref{sec:approach}.
The following research questions guide our experimental study:
\begin{inparaenum}[\bf {\bf RQ}1\upshape)]
\item What is the impact of applying \textit{EABlock} in KG creation in terms of execution time?
\item How does applying \textit{EABlock} in the process of KG creation impact the quality of the result KG?
\item How sensitive to the quality of input data is \textit{EABlock}? 
\end{inparaenum}
As a proof of concept, we set up the experiments using biomedical data. Accordingly, we rely on an API of Falcon\furl{https://labs.tib.eu/sdm/biofalcon/} that provides a filtered subset of the BK~\cite{Sakor2020} omitting the resources that are not related to the biomedical domain. 

\subsection{\textit{EABlock} Efficiency- RQ1}
To evaluate how the performance of a KG creation pipeline may be impacted applying EABlock, we set up 24 KG creation pipelines in overall. Experiments are grouped as Baseline or \textit{EABlock}; Baseline corresponds to the pipelines where execution of EA is in a pre-processing stage, while \textit{EABlock} represent the KG creation pipelines in which \textit{EABlock} enables the specification of EA in the RML mapping rules. Experiments are grouped into six categories, each category utilizing a different $DE$, i.e., all the experiments in one category have the same $DE$. To avoid any bias caused by the techniques applied in the development of the state-of-the-art engines, we repeat the same experiments by two different available engines including RocketRML~\furl{https://github.com/semantifyit/RocketRML} and SDM-RDFizer~\furl{https://github.com/SDM-TIB/SDM-RDFizer}. Accordingly, the experiments in one category differs in \textbf{a.} the applied RML-compliant engine and \textbf{b.} whether \textit{EABlock} is used as part of the pipeline or not. 
\noindent\textbf{Datasets and Mappings.} Considering the parameters that affect the performance of KG creation pipelines \cite{chaves2019parameters}, we define three different sets of mapping rules, which are distinguished based on the complexity of the \texttt{rr:TriplesMap}s that refers to the \textit{EABlock} transformation functions. We manipulate the complexity of the mentioned rules by having different number of \texttt{rr:RefObjectMap}s, i.e., zero, one, or two \texttt{rr:RefObjectMap} (referred to as noROM, 1ROM, and 2ROM respectively, in Figure~\ref{fig:efficiency}). In an attempt to prevent possible effects of data volume on the results of the experiments, we generate two relatively small datasets including 1,000 and 2,000 randomly selected records. Each dataset comprises 22 attributes, two of which are referenced in the mapping rules. \textbf{Setups.} We define two KG creation pipelines, \textit{Baseline} and \textit{EABlock}, which execute the same entity alignment tasks and produce the same KG. 
The \textit{Baseline} pipeline evaluates RML mapping rules while the entity alignment is performed in a pre-processing step.  
Contrary, the \textit{EABlock} pipeline encapsulates these tasks in the \textit{EABlock} functions that are called in the RML mapping rules. \textbf{Metrics.} 
\textit{Execution time:} Elapsed time spent by the whole pipeline to complete the creation of a KG; it is measured as the absolute wall-clock system time as reported by the \texttt{time} command of the Linux operating system. The experiments were run in an Intel(R) Xeon(R) equipped with a CPU E5-2603 v3 @ 1.60GHz 20 cores, 64GB memory and with the O.S. Ubuntu 16.04LTS. \textbf{Results.} \autoref{fig:efficiency} illustrates the performance of two approaches of KG creations i.e., Baseline which perform the EA as pre-processing, and \textit{EABlock} which enables the specification of EA as part of the RML mapping rules. As it can be observed in \autoref{fig:efficiency}, independent of the applied RML-compliant engine utilizing \textit{EABlock} in all KG pipelines reduces the overall execution time of the KG creation. \autoref{fig:efficiency} demonstrates that performing EA as pre-processing is more expensive than using \textit{EABlock} as part of the main pipeline of KG creations. It can also be observed that in case of having more complex mapping rules, the impact of \textit{EABlock} in decreasing the execution time is even more considerable and significant. 
\raggedbottom

\raggedbottom
\subsection{\textit{EABlock} Effectiveness - RQ2}
\begin{figure*}[t!]
 \centering
        \begin{subfigure}{.33\textwidth}
        \includegraphics[trim=0 0 0 21,clip,width=\linewidth]{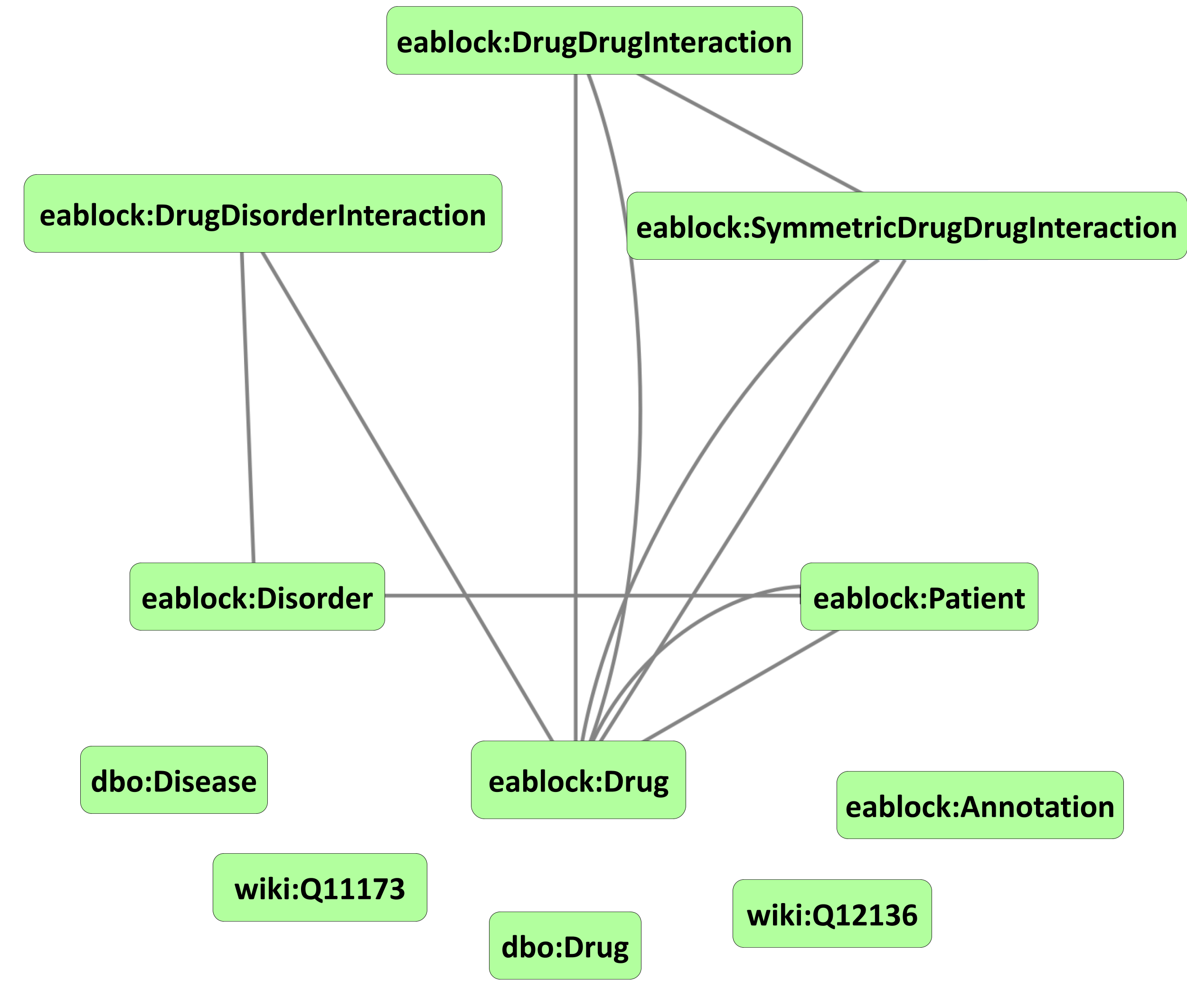}
         \caption{Directed Labelled Graph $G_b$}
            \label{fig:baseline}
       \end{subfigure}~
        \begin{subfigure}{.33\textwidth}     
        \includegraphics[trim=0 0 0 21,clip,width=\linewidth]{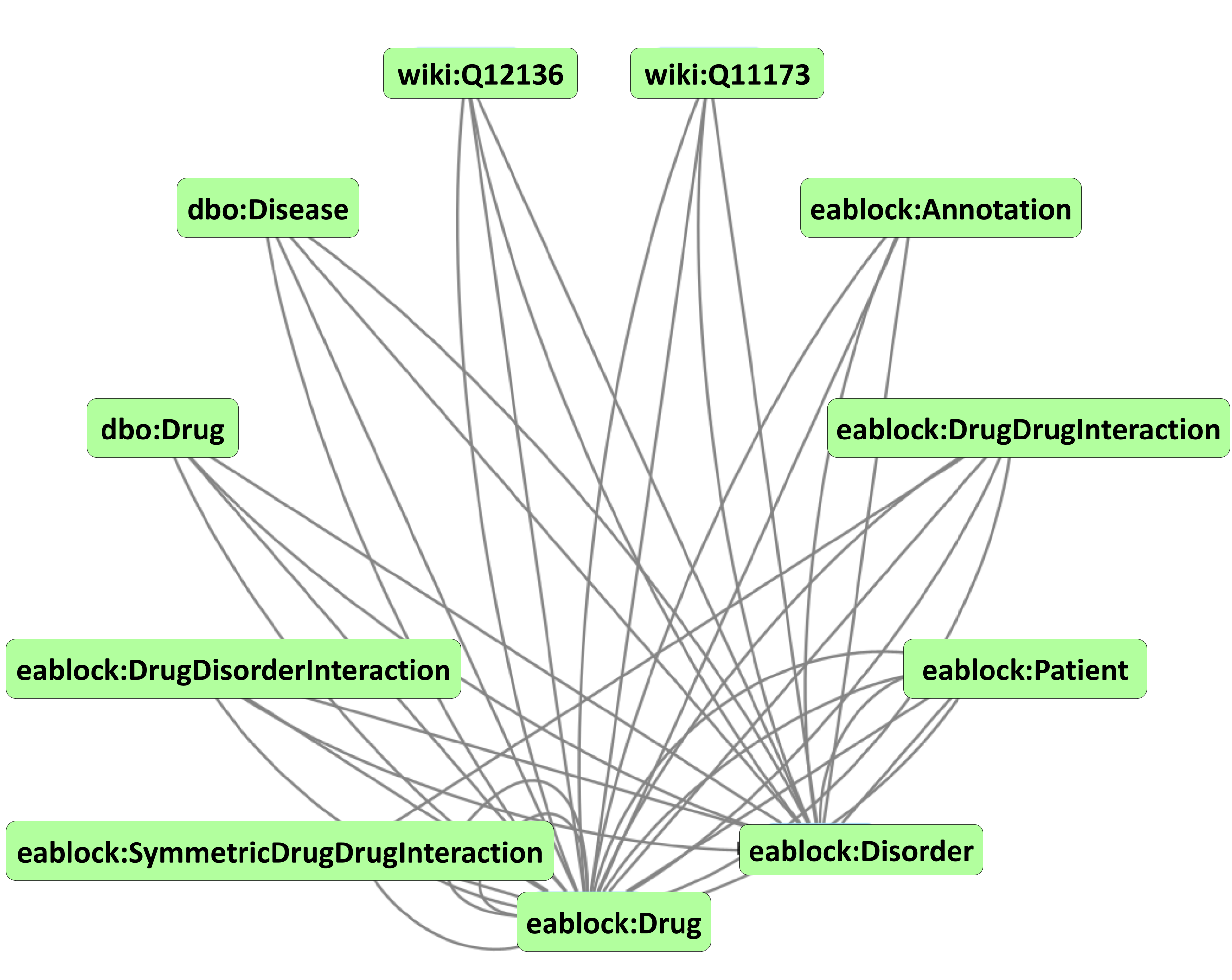}
        \caption{Directed Labelled Graph $G_{eablock}$}
            \label{fig:EABlock}
         \end{subfigure}~
         \begin{subfigure}{.30\textwidth}     
\resizebox{1.0\linewidth}{!}{%
\begin{tabular}{|l|c|c|}
\hline
\rowcolor[HTML]{98C6B9} 
\textbf{Analysis}          & \multicolumn{1}{l|}{\cellcolor[HTML]{98C6B9}\textbf{EABlock} ($G_{eablock}$)} & \multicolumn{1}{l|}{\cellcolor[HTML]{98C6B9}\textbf{Baseline} ($G_{b}$)} \\ \hline
\rowcolor[HTML]{D4FACA} 
Number of nodes            & 11                                                              & 11                                                             \\ \hline
\rowcolor[HTML]{D4FACA} 
Number of edges            & 39                                                              & 10                                                             \\ \hline
\rowcolor[HTML]{D4FACA} 
Avg. number of neighbors   & 3.091                                                           & 1.273                                                          \\ \hline
\rowcolor[HTML]{D4FACA} 
Network diameter           & 2                                                               & 1                                                              \\ \hline
\rowcolor[HTML]{D4FACA} 
Clustering coefficient     & 0.184                                                           & 0.098                                                          \\ \hline
\rowcolor[HTML]{D4FACA} 
Network density            & 0.155                                                           & 0.064                                                          \\ \hline
\rowcolor[HTML]{D4FACA} 
Connected components       & 1                                                               & 6                                                              \\ \hline
\end{tabular}
}
        \caption{Graph metrics for $G_{b}$ and $G_{eablock}$}
            \label{tab:connectivity}
         \end{subfigure}
\caption{\textbf{Connectivity Analysis}. Baseline and \textit{EABlock} pipelines generate $KG_b$ and $KG_{eablock}$, respectively. $KG_b$ and $KG_{eablock}$ have the same classes and entities. However, $KG_{b}$ does not include the entity alignments to UMLS, Wikidata, and DBpedia added to $KG_{eablock}$. $G_b$ and $G_{eablock}$ are directed labelled graphs that provide an aggregated representation of $KG_b$ and $KG_{eablock}$. The values of the graph metrics corroborate that connectivity is increased by entity alignment performed by the \textit{EABlock} pipeline.}
    \label{fig:networkanalysis}
\end{figure*}

We define two pipelines: \textit{Baseline} and \textit{EABlock}; the \textit{Baseline} pipeline includes no entity alignment task. The aim is to evaluate the connectivity in a KG created using the \textit{EABlock} pipeline and assess RQ2.
\noindent\textbf{Datasets and Mappings.} We extract data related to drugs (11,293 records), the disorders for which the drugs are prescribed (416 records), and the interactions between the drugs (1,646,836 records) from DrugBank\furl{https://go.drugbank.com/} (version 5.1.8). We produce three mock datasets resembling normal clinical notes for cancer patients, including the data related to the comorbidities (1,322 records) and prescribed oncological (1,764 records) and non-oncological drugs (1,325 records). We create a unified schema for these datasets and a set $D_4$ of RML mappings rules to integrate them. Also, we create a set $D_5$ with all the mapping rules in, $D_4$ plus the corresponding calls to \textit{EABlock} functions to execute entity alignment for drugs and disorders. 
\noindent\textbf{Analysis.}
Let $KG_b$ and $KG_{eablock}$ be the KGs created by the \textit{Baseline} and \textit{EABlock} pipelines, respectively. $KG_{eablock}$
comprises 10,339,870 RDF triples, while $KG_b$ has 10,200,209. 
$KG_b$ and $KG_{eablock}$ 
 are used to create two labelled directed graphs $G_b=(V,E_b)$ and $G_{eablock}=(V,E_{eablock})$ and traditional network analysis methods are applied to determine connectivity. Vertices in $V$ keeps the classes in $KG_b$ and $KG_{eablock}$ with at least one resource; $KG_b$ and $KG_{eablock}$ have the same resources and literals. A labelled directed edge $e=(q,p,k)$ belongs to $E_b$ (resp. to $E_{eablock}$) if there are classes $Q$ and $K$ in $V$, and $q$ and $k$ are instances of $Q$ and $K$ in $KG_b$ (resp., $KG_{eablock}$) and the RDF triple $(q p k)$ belongs to $KG_b$ (resp., $KG_{eablock}$). $G_b$ and $G_{eablock}$ provide an aggregated representation of $KG_b$ and $KG_{eablock}$. \autoref{fig:networkanalysis} depicts $G_b$ and $G_{eablock}$; $G_{eablock}$ is composed of 11 vertices and 39 directed edges. While, $G_b$ comprises 11 vertices connected by only 10 edges. Table \ref{tab:connectivity} compares $G_b$ and $G_{eablock}$ in terms of graph metrics generated by Cytoscape\furl{https://cytoscape.org/}. \textbf{Metrics.} \textbf{a. }Average number of neighbors indicates the average connectivity of a vertex or node in a graph. \textbf{b.} Network diameter measures the shortest path that connects the two most distant nodes in a graph. \textbf{c.} Clustering coefficient measures the tendency of nodes who share the same connections in a graph to become connected. If a neighborhood is fully connected, the clustering coefficient is 1.0 while a value close to 0.0 means that there is no connection in the neighborhood. \textbf{d.} Network density measures the portion of potential edges in a graph that are actually edges; a value close to 1.0 indicates that the graph is fully connected. \textbf{e. }The number of connected components indicates the number of subgraphs composed of vertices connected by at least one path. \textbf{Results.}
The results reported in \autoref{tab:connectivity} indicate the average number of neighbors in $KG_{eablock}$ comprises entities that are more connected. Moreover, the clustering coefficient is relatively low, but the CUIs annotations and links to DBpedia and Wikidata included in $KG_{eablock}$,  increase the connectivity in the neighborhoods of $G_{eablock}$. In particular,  \texttt{eablock:Patient}, \texttt{eablock:DrugDisorder}, \texttt{eablock:Annotation}, \texttt{eablock:Disea}- \texttt{se}, \texttt{wiki:Q12136}, \texttt{wiki:Q11173}, and \texttt{dbo:Drug} have a neighborhood connectivity of 8 in $G_{eablock}$. On the other hand, in $G_b$, the neighborhood connectivity of \texttt{eablock:Annotation}, \texttt{eablock:Di}- \texttt{sease}, \texttt{wiki:Q12136}, \texttt{wiki:Q11173}, and \texttt{dbo:Drug} is 0, and   
\texttt{eabloc}- \texttt{k:Patient} and \texttt{eablock:DrugDis}-\texttt{orderInteraction} is 3. These results corroborate that connectivity is enhanced as a result of the entity alignment implemented by the \textit{EABlock} functions.

\subsection{\textit{EABlock} Effectiveness - RQ3}
We evaluate the performance of the \textit{EABlock} and the impact that data quality issues may have on entity alignment. We define 15 tesbeds, five for DBpedia, for Wikidata, and for UMLS. 
\noindent\textbf{Datasets} \textit{Gold standard:} We create the gold standard datasets by extracting biomedical instances from DBpedia (06-2010) and Wikidata (04-2019) KGs, and the UMLS (November 2020) dataset. We obtain 51,850 records from Wikidata considering the classes ``Q12136 (disease)", ``Q3736076 (biological function)", ``Q43229 (organization)", ``Q514 (anatomy)", ``Q79529 (chemical substance)", and ``Q863908 (nucleic acid sequence)". From DBpedia, we extract 43,680 records related to the classes ``ChemicalSubstance", ``Disease", ``Gene", and ``PersonFunction". We consider the whole dataset of UMLS including 3,741,395 records. However, to avoid having a huge difference between the number of records with the experiments of DBpedia and Wikidata, we randomly select 1,496,557 records from the UMLS dataset. Meanwhile, to decrease bias, for each of the five experiments, we generate a new randomly selected dataset of UMLS with the same number of records, i.e., 1,496,557\furl{https://tib.eu/cloud/s/XJiqDDAHqM8Fw5K}. \textbf{Testbeds:} Five testbeds are generated by manipulating the gold standard datasets considering frequent quality issues that may exist in datasets; character capitalization, elimination, insertion, and replacement. For this purpose, for DBpedia, Wikidata, and UMLS, we create five datasets from the related gold standard dataset, but with lower quality than the gold standard. To decrease the quality of the data, we intentionally introduce misspelling errors in values of the records. The errors include: \begin{inparaenum}[\bf {\bf}a\upshape)]
\item capitalizing all the characters of a record value;
\item randomly eliminating one character from the value;
\item randomly replacing one character with another randomly selected character; and
\item inserting a randomly selected character to a random location in the record value.
\end{inparaenum}
Accordingly, each of mentioned errors are introduced in 50\% of the records in one of the testbeds, the other 50\% of the records carry the same values as the gold standards. The last testbed created by including all four types of errors, has the lowest quality. In this dataset, each 20\% out of 80\% of the records involves exactly one of the four errors. Therefore, 20\% of the records are error-free, in contrast to the other four test beds, in which 50\% of all records are free from errors. \textbf{Metrics.}
We assess the sensitivity of \textit{EABlock} in terms of \textbf{I.} precision; the fraction of correct results of entity alignments from all the entity alignment results returned by \textit{EABlock}, \textbf{II.} recall; the fraction of correct results of performed entity alignment from the all expected results to be returned by \textit{EABlock}, \textbf{III.} F1 score; the harmonic mean of precision and recall. \textbf{Results.}
\autoref{tab:quality} demonstrates the results of running \textit{EABlock} over the five configurations of error types explained in \textit{Testbeds}. The entity alignment engine (Falcon) reports relatively high performance. Cleaning the BK of Falcon and filtering it to contain only resources related to the biomedical domain, as explained in \autoref{sec:approach}, reduces the ambiguity among the resources in the BK. This plays a major role in having such high performance. Also, having the input of the entity alignment module as keywords without any noise, e.g., stopwords, helps the module recognize and link the labels precisely. Moreover, \autoref{tab:quality} suggests that the used entity alignment module is able to overcome the proposed error types. There are records for which \textit{EABlock} fails to return the expected linked entity based on the gold standard. These failures are mostly caused by reasons other than the introduced error types; we enumerate a couple of these examples with possible explanations:
\begin{inparaenum}[\bf a\upshape)]
\item there are cases that the retrieved linked entities are correct, although their identifiers differ from those in the gold standard. For instance, for the keyword ``malignant histiocytosis" \textit{EABlock} retrieves ``Q164952"\furl{https://www.wikidata.org/wiki/Q164952} from Wikidata, while, in the gold standard the Wikidata identifier for the same keyword is ``Q52962465"\furl{http://www.wikidata.org/entity/Q52962465}. However, both identifiers lead to the same entry in Wikidata; for some keywords, more than one identifier exists in such KGs.
\item Another example of unexpected retrieved linked entities, can be observed in case of having long combinations of keywords, such as ``early infantile epileptic encephalopathy 19". In this case, \textit{EABlock} links the first entity that can be recognized by the first couple of keywords; ``Q61913448"\furl{https://www.wikidata.org/wiki/Q61913448} which belongs to the label ``early infantile epileptic encephalopathy 37". The same failure case can be observed in retrieval from DBpedia as well; for the keyword ``Chronic leukemia" \textit{EABlock} retrieves a link to ``B-cell\_chronic\_lymphocytic\_leukemia" \furl{https://dbpedia.org/page/Chronic_lymphocytic_leukemia} while based on the gold standard the correct link is ``Chronic\_leukemia"\furl{https://dbpedia.org/page/Chronic_leukemia}. 

\end{inparaenum}   

\subsection{Disscusion}
As is emphasized in previous sections, \textit{EABlock} can be utilized in any KG creation pipeline that applies an RML-compliant engine for the tasks of mapping rules translation and RDF triples generation. Although there is a W3C community to tackle KG creation using R2RML and RML mapping languages \furl{https://www.w3.org/community/kg-construct/}, there is a lack of formal definitions, which is a barrier and impacts on general adoption of these engines. Another aspect of being agnostic is the application of a tool that can perform the NER and EL tasks. Despite all the values this independence brings, the performance of \textit{EABlock} is subject to change based on the choice of the entity alignment tool.      

\begin{table}[ht!]
\scriptsize
\centering
\caption{\textit{Effectiveness}. The \textit{EABlock} performance is assessed in 15 datasets. The aligned entities are compared with the resources of the original labels.}
\label{tab:quality}
\resizebox{0.8\linewidth}{!}{%
\begin{tabular}{lccl}
\hline
\rowcolor[HTML]{98C6B9} 
\multicolumn{4}{c}{\cellcolor[HTML]{98C6B9}\textbf{UMLS}}                                                                                                                                                                                                                        \\ \hline
\rowcolor[HTML]{B3D3A5} 
\multicolumn{1}{|l|}{\cellcolor[HTML]{fbff12}\textbf{Text Error Type}}         & \multicolumn{1}{c|}{\cellcolor[HTML]{fbff12}\textbf{Precision}} & \multicolumn{1}{c|}{\cellcolor[HTML]{fbff12}\textbf{Recall}} & \multicolumn{1}{l|}{\cellcolor[HTML]{fbff12}\textbf{F1 Score}} \\ \hline
\rowcolor[HTML]{E3FFDC} 
\multicolumn{1}{|l|}{\cellcolor[HTML]{E3FFDC}Capitalization of all characters} & \multicolumn{1}{c|}{\cellcolor[HTML]{E3FFDC}1.0}       & \multicolumn{1}{c|}{\cellcolor[HTML]{E3FFDC}0.97}   & \multicolumn{1}{l|}{\cellcolor[HTML]{E3FFDC}0.99}     \\ \hline
\rowcolor[HTML]{E3FFDC} 
\multicolumn{1}{|l|}{\cellcolor[HTML]{E3FFDC}Elimination of a character}       & \multicolumn{1}{c|}{\cellcolor[HTML]{E3FFDC}0.99}               & \multicolumn{1}{c|}{\cellcolor[HTML]{E3FFDC}0.74}            & \multicolumn{1}{l|}{\cellcolor[HTML]{E3FFDC}0.85}              \\ \hline
\rowcolor[HTML]{E3FFDC} 
\multicolumn{1}{|l|}{\cellcolor[HTML]{E3FFDC}Replacement of a character}       & \multicolumn{1}{c|}{\cellcolor[HTML]{E3FFDC}0.99}               & \multicolumn{1}{c|}{\cellcolor[HTML]{E3FFDC}0.75}            & \multicolumn{1}{l|}{\cellcolor[HTML]{E3FFDC}0.85}              \\ \hline
\rowcolor[HTML]{E3FFDC} 
\multicolumn{1}{|l|}{\cellcolor[HTML]{E3FFDC}Insertion of a new character}     & \multicolumn{1}{c|}{\cellcolor[HTML]{E3FFDC}0.99}               & \multicolumn{1}{c|}{\cellcolor[HTML]{E3FFDC}0.50}            & \multicolumn{1}{l|}{\cellcolor[HTML]{E3FFDC}0.66}              \\ \hline
\rowcolor[HTML]{E3FFDC} 
\multicolumn{1}{|l|}{\cellcolor[HTML]{E3FFDC}Combination of all 4 errors}      & \multicolumn{1}{c|}{\cellcolor[HTML]{E3FFDC}1.0}       & \multicolumn{1}{c|}{\cellcolor[HTML]{E3FFDC}0.78}            & \multicolumn{1}{l|}{\cellcolor[HTML]{E3FFDC}0.88}              \\ \hline
\rowcolor[HTML]{98C6B9} 
\multicolumn{4}{c}{\cellcolor[HTML]{98C6B9}\textbf{DBpedia}}                                                                                                                                                                                                                     \\ \hline
\rowcolor[HTML]{B3D3A5} 
\multicolumn{1}{|l|}{\cellcolor[HTML]{fbff12}\textbf{Text Error Type}}         & \multicolumn{1}{c|}{\cellcolor[HTML]{fbff12}\textbf{Precision}} & \multicolumn{1}{c|}{\cellcolor[HTML]{fbff12}\textbf{Recall}} & \multicolumn{1}{l|}{\cellcolor[HTML]{fbff12}\textbf{F1 Score}} \\ \hline
\rowcolor[HTML]{E3FFDC} 
\multicolumn{1}{|l|}{\cellcolor[HTML]{E3FFDC}Capitalization of all characters} & \multicolumn{1}{c|}{\cellcolor[HTML]{E3FFDC}0.78}      & \multicolumn{1}{c|}{\cellcolor[HTML]{E3FFDC}0.78}   & \multicolumn{1}{l|}{\cellcolor[HTML]{E3FFDC}0.78}     \\ \hline
\rowcolor[HTML]{E3FFDC} 
\multicolumn{1}{|l|}{\cellcolor[HTML]{E3FFDC}Elimination of a character}       & \multicolumn{1}{c|}{\cellcolor[HTML]{E3FFDC}0.78}               & \multicolumn{1}{c|}{\cellcolor[HTML]{E3FFDC}0.78}   & \multicolumn{1}{l|}{\cellcolor[HTML]{E3FFDC}0.78}              \\ \hline
\rowcolor[HTML]{E3FFDC} 
\multicolumn{1}{|l|}{\cellcolor[HTML]{E3FFDC}Replacement of a character}       & \multicolumn{1}{c|}{\cellcolor[HTML]{E3FFDC}0.98}               & \multicolumn{1}{c|}{\cellcolor[HTML]{E3FFDC}0.98}   & \multicolumn{1}{l|}{\cellcolor[HTML]{E3FFDC}0.98}              \\ \hline
\rowcolor[HTML]{E3FFDC} 
\multicolumn{1}{|l|}{\cellcolor[HTML]{E3FFDC}Insertion of a new character}     & \multicolumn{1}{c|}{\cellcolor[HTML]{E3FFDC}0.78}               & \multicolumn{1}{c|}{\cellcolor[HTML]{E3FFDC}0.78}            & \multicolumn{1}{l|}{\cellcolor[HTML]{E3FFDC}0.78}              \\ \hline
\rowcolor[HTML]{E3FFDC} 
\multicolumn{1}{|l|}{\cellcolor[HTML]{E3FFDC}Combination of all 4 errors}      & \multicolumn{1}{c|}{\cellcolor[HTML]{E3FFDC}0.78}               & \multicolumn{1}{c|}{\cellcolor[HTML]{E3FFDC}0.78}            & \multicolumn{1}{l|}{\cellcolor[HTML]{E3FFDC}0.78}              \\ \hline
\rowcolor[HTML]{98C6B9} 
\multicolumn{4}{c}{\cellcolor[HTML]{98C6B9}\textbf{Wikidata}}                                                                                                                                                                                                                    \\ \hline
\rowcolor[HTML]{B3D3A5} 
\multicolumn{1}{|l|}{\cellcolor[HTML]{fbff12}\textbf{Text Error Type}}         & \multicolumn{1}{c|}{\cellcolor[HTML]{fbff12}\textbf{Precision}} & \multicolumn{1}{c|}{\cellcolor[HTML]{fbff12}\textbf{Recall}} & \multicolumn{1}{l|}{\cellcolor[HTML]{fbff12}\textbf{F1 Score}} \\ \hline
\rowcolor[HTML]{E3FFDC} 
\multicolumn{1}{|l|}{\cellcolor[HTML]{E3FFDC}Capitalization of all characters} & \multicolumn{1}{c|}{\cellcolor[HTML]{E3FFDC}0.99}      & \multicolumn{1}{c|}{\cellcolor[HTML]{E3FFDC}0.99}            & \multicolumn{1}{l|}{\cellcolor[HTML]{E3FFDC}0.99}              \\ \hline
\rowcolor[HTML]{E3FFDC} 
\multicolumn{1}{|l|}{\cellcolor[HTML]{E3FFDC}Elimination of a character}       & \multicolumn{1}{c|}{\cellcolor[HTML]{E3FFDC}0.99}      & \multicolumn{1}{c|}{\cellcolor[HTML]{E3FFDC}0.99}            & \multicolumn{1}{l|}{\cellcolor[HTML]{E3FFDC}0.99}              \\ \hline
\rowcolor[HTML]{E3FFDC} 
\multicolumn{1}{|l|}{\cellcolor[HTML]{E3FFDC}Replacement of a character}       & \multicolumn{1}{c|}{\cellcolor[HTML]{E3FFDC}0.99}      & \multicolumn{1}{c|}{\cellcolor[HTML]{E3FFDC}0.99}            & \multicolumn{1}{l|}{\cellcolor[HTML]{E3FFDC}0.99}              \\ \hline
\rowcolor[HTML]{E3FFDC} 
\multicolumn{1}{|l|}{\cellcolor[HTML]{E3FFDC}Insertion of a new character}     & \multicolumn{1}{c|}{\cellcolor[HTML]{E3FFDC}0.99}      & \multicolumn{1}{c|}{\cellcolor[HTML]{E3FFDC}0.99}            & \multicolumn{1}{l|}{\cellcolor[HTML]{E3FFDC}0.99}              \\ \hline
\rowcolor[HTML]{E3FFDC} 
\multicolumn{1}{|l|}{\cellcolor[HTML]{E3FFDC}Combination of all 4 errors}      & \multicolumn{1}{c|}{\cellcolor[HTML]{E3FFDC}0.99}      & \multicolumn{1}{c|}{\cellcolor[HTML]{E3FFDC}0.99}   & \multicolumn{1}{l|}{\cellcolor[HTML]{E3FFDC}0.99}     \\ \hline
\end{tabular}
}
\end{table}

\raggedbottom

\section{Conclusions and Future Work}
\label{sec:conclusion}
The diverse interoperability issues existing in textual data and the demand of having a transparent, traceable, and efficient pipeline of KG creation led us to introduce \textit{EABlock}. \textit{EABlock} is an approach to solve entity alignment problems by capturing knowledge from existing KGs while keeping the procedure transparent and traceable. With an eager evaluation strategy and efficient translation of mapping rules into function-free rules, \textit{EABlock} ensures not to sacrifice the efficiency at the cost of reproducibility. The observed experimental results show the benefits of grounding solutions for KG creation in the well-established problems like NER, NL, and data integration systems. Thus, \textit{EABlock} broadens the repertory of approaches for KG creation and provides the basis for developing real-world KGs. Our vision is that \textit{EABlock} will be the starting point for the development of FnO functions which can be made available following the FAIR principles. 
In the future, we will empower the \textit{EABlock} with a set of functions to overcome interoperability problems across multilingual and unstructured datasets.

\noindent\paragraph{{\textbf{Acknowledgments.}}} 
\noindent This work has been partially supported by the EU H2020 RIA
funded project CLARIFY with grant agreement No 875160 and P4-LUCAT No 53000015.

\bibliographystyle{abbrv}
\bibliography{biblio}
\end{document}